# TEXTBOOK TO TRIPLES: Creating knowledge graph in the form of triples from AI TextBook

Aman Kumar
Edward P. Fitts Department of Industrial and Systems
Engineering, NC State University
Raleigh, United States
akumar33@ncsu.edu

Swathi Dinakaran
Department of Computer Science and Engineering,
NC State University
Raleigh, United States
sdinaka@ncsu.edu

*Abstract*—A knowledge graph is an essential and trending technology with great applications in entity recognition, search, or question answering. There are a plethora of methods in natural language processing for performing the task of Named entity recognition; however, there are very few methods that could provide triples for a domain-specific text. In this paper, an effort has been made towards developing a system that could convert the text from a given textbook into triples that can be used to visualize as a knowledge graph and use for further applications. The initial assessment and evaluation gave promising results with an F1 score of 82%.

## I. INTRODUCTION

Information extraction (IE) is the task of extracting structured knowledge automatically from unstructured and/or semi-structured machine-readable documents and from other electronically represented sources, and IE further can be used to develop Knowledge graphs (KG). There are many applications of Knowledge graphs in the domain, such as search [1], question answering [2], and recommendation system [3]. Throughout the past work, multiple knowledge graph databases in the form of triples have been introduced, such as Wikidata, YAGO, Satori, etc. The wikidata [4] triples have been generated based on the Wikipedia website and are developed by the contribution of volunteers through crowdsourcing. Any system that could assist in constructing the domain-specific knowledge graph, which requires the knowledge/expertise in the area, would accelerate the process of KG development and will allow the creation of any new/obscure domain KG.

It is unsurprising that state-of-the-art performances appear to be related to web search, considering that knowledge graphs were initially developed to boost search engines. The Web can be converted from a series of hyperlinks into a set of concept links via a knowledge graph. This implies that the users would be able to retrieve information about subjects based on true semantics and not simply by matching characters.

For years, the creation of information graphs from text has been a difficult issue [5]. A pipeline of NLP functions, such as named entity recognition followed by coreference resolution, entity linking, and then relationship extraction, is the prevalent method [6], [7]. It is very important to have a proper segmentation and understanding of the concept of knowledge graph and data graph. Most of the existing 'KG' are data graphs that are not capable of extracting the knowledge but are only able to represent the data in the form of text that is present in a sentence. Therefore the concept of coreference plays a vital role in identifying the knowledge and converting the data graph into a knowledge graph.

There are three components of the knowledge graph construction process, which include information extraction, knowledge fusion, and knowledge processing. For the extraction of information, the components of triples are entity and relations. An entity is also known as a subject or object, and the relation between the entities is also termed as a predicate.

## II. MOTIVATING EXAMPLE:

For any domain-specific textbook, there is a lot of information available that requires patience to study and gain knowledge. This becomes even more difficult when someone from a different domain is trying to understand the concept.

The motivation is to focus on the relationship extraction part of knowledge graph building since there are no relationship extraction methods known that work promisingly well for any domain-specific text. E.g., The unsupervised OpenIE method for relation extraction doesn't give a promising relation between subject and predicate. Given predicate and object, OpenIE by Allen institute queries the subject that means it is asking users about the entities. Most of the OpenIE, including Stanford's one, doesn't take into account domain-specific entities. Moreover, the dependency parsing method couldn't know about an entity of an unknown domain unless custom NER has been done.

For example, the case when the user is presented with the text from a textbook that is non-domain specific, there is no off-the-shelf methodology to extract triples and possibly form a knowledge graph with the input of only the text from the textbook. Since the other methods of extraction require the use of ontology, domain-specific knowledge, etc., to successfully identify the entities and relations present in the textbook, we are proposing a novel methodology that takes two inputs from the text of the textbook:
1. The index or glossary of terms
2. The text that needs to be converted into a knowledge graph.

For example, in the textbook on Artificial Intelligence [13] that we have taken, the following terms are present in the index :

*A. Input:*

*Terms in glossary*:
Search
Multiple sequences of moves
known-state
*Given the sentence, that needs to be created as a knowledge graph* :
Search is where multiple sequences of moves may lead to the known-state

*B. Main Output:*

$$[search, is, multiple sequences of moves]$$

$$[multiple sequences of moves, lead to, known\text{-}state]]$$

In the given example, where the inputs are terms in the glossary and the sentence that needs to be used to create a knowledge graph.
Our approach uses methods to extract entities and performs chunking on the extracted entities and the remainder of the text, to get the entities and relations. Our method also does processing on the text to extract entities and relations to combine these as a triple.

*C. Output 2:*

Now with the triples extracted in place, we form a graph and visualize the extracted triples in the form of a knowledge graph. The knowledge graph has triples in the following form: node as a subject, node as an object, edge label as relation name, edge direction from subject to object. For example,
Text:
An Agent has sensors.
It has actuators too.
An agent gives output through Actuators.
Triples:

$$[agent, has, sensors]$$

$$[agent, has, actuators]$$

$$[agent, gives, output]$$

$$[output, through, sensors]$$

*D. Output 3:*

Another useful insight from the knowledge graph is: given a text, find all the relations that a single entity gets involved in. For example,
Text:
An Agent has sensors.
Some text
Some text
An agent gives output through Actuators.
If the user wants to know the context in which the term agent occurs, then we form a knowledge graph from only the triples that has agent involved.
triples used in forming knowledge graph:

$$[agent, has, sensors]$$

$$[agent, gives, output]$$

## III. RELATED WORK

Some of the previous works involved in forming a knowledge graph out of a textbook involved adapting the method in accordance with the particular domain. Moreover, such domain-specific methods often make use of an ontology or knowledge base that can be used as a basis while extracting entities and relations. In [8], with the use of Long Short-Term Memory networks [lstm, mapTextKG, embeddingModelsEntities], text can also be translated to knowledge graph entities. To construct skip gram entities that are included in a knowledge base, supervised learning by the use of Random Walk and an LSTM recurrent network is used. Based on semantic similarity, the input text is then assigned to these entities. To train the models, paired samples of text and knowledge graph entities need this.

But these methods have the limitation of the need for extra knowledge in terms of either an existing knowledge base or an enormous training set. Nonetheless, when it comes to real-world settings, such as the domain-specific text in rare domains, we face the problem of low-resource data similar to public security—translating between uncommon languages by computer. There's no annotated dataset benchmark applicable to certain domains, and it is almost impossible to find the correct pivot language to take advantage of the latest high-resource NER instruments. Hence we propose a simple triple extraction mechanism that extracts triples without any extra data or knowledge required. The method uses the knowledge gained from the entities present in the glossary and the given in the textbook itself.

Another simpler implementation [9] involves systems built using well established natural language processing frameworks such as NLTK and SpaCy, and makes use of standard techniques such as tokenization, part-of-speech (POS) tagging, named entity recognition, coreference resolution, and noun/verb phrase chunking.

The above work doesn't involve a textbook and works well for general text. However, the limitations are the redundancy in the triples generated. Hence we adopted the above-mentioned method to include domain specificity by using the keywords from the glossary. We incorporated several of our own algorithms while dealing with the problem. Additionally, for the convenience of entities visualization for understanding a particular text, we have introduced visualization methods by plotting the above in the form of a knowledge graph.

## IV. METHODOLOGY

Representation of the data is the most difficult task taking into consideration the fact that a phrase could have zero or multiple triples. In order for the relationships to be derived directly from the text, it becomes very difficult for the representation of the input data in a way that a decently sized set of valid triples from a given document could be predicted by the model. We chose not to take a deep learning method approach such as snorkel [12] as it is required for this approach to have some relations prior in order to train it for detecting further relations, and hence we draw on the wide range of NLP resources readily available. Figure 1 shows a flowchart showing the pipeline for creating triples from the text, that is explained below:

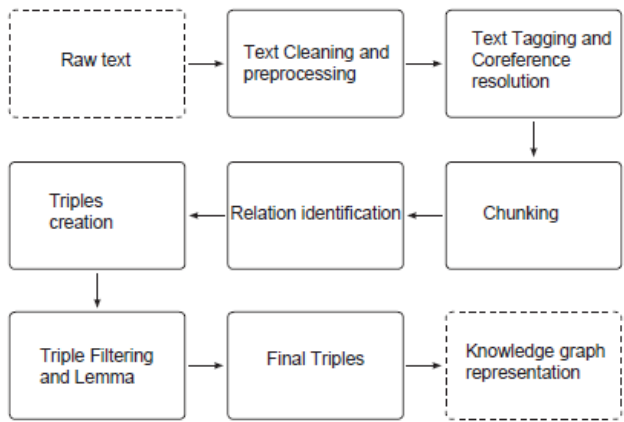

Fig. 1. Flow diagram for conversion of text to KG.

### A. Cleaning of text taken from the textbook:

Since we are focussed on a single chapter of a domain-specific text, the data cleaning part can be simplified by selecting only the page-wise content for a specific chapter in the textbook, by converting the text from pdf to doc and taking only the relevant contents or using our previous work on text cleaning for thesis related work.

(i). The entities can be identified from the words given at the index at the end of the book, as shown above. (ii). The keywords present in the page range of the chapter chosen is filtered as the keyword list. (iii). These keywords should then be identified from within the chapter and can be represented as entities.

### B. Spacy methods for text tagging:

The text from a chapter, here chapter 2, in the form of a few paragraphs, is the major text that needs to be extracted in the form of a knowledge graph or triples. The keyword list extracted from the previous step is used for tagging the above text. These keywords should then be identified from within the chapter and can be represented as entities. Further, spacy is used for tagging the text using spacy , which returns dataframes with the following fields: ['Document', 'Sentence', 'Word', 'Word', 'EntityType', 'EntityIOB', 'Lemma', 'POS', 'POSTag', 'Start', 'End', 'Dependency']

### C. Spacy neuralcoref for Coreference resolution

The sentences containing coreference pointers are resolved by the addition of neural coref along with the SpaCy that's already present in the pipeline.
E.g., for the sentence,
'a rational agent should select an action that is expected to maximize its performance measure'
Triples extracted from the output is given below:

$$[rational\ agent, should\ select, action]$$
$$[action, is expected to maximize, performance measure]$$

### D. Manually crafted methods for chunking of text

The entities are marked after coreferencing and chunking. The chunking of text involves grouping individual words into phrases by using certain general chunking rules. The following pseudocode explains the chunking rules used by our method

---
**Algorithm 1** procedure CHUNKPHRASES(document)
   **for** each sentence in document **do**
      Chunk noun phrases (NPs) and tag as ENTITY
      chunk NPs
      chunk ( NP )
      chunk NP + of + NP
      chunk NP + NP
      chunk verb + particle
      chunk verb + adpositions
      chunk adpositions + verb
      chunk particle + verb
      chunk verb + verb
   **end for**
   return document

---

### E. Relation identification

All the verbs that are extracted from the given text are stored in an array, as possible relation candidates or predicates. Further, the words immediately after the verbs are also taken as adposition predicates.

### F. Triples creation

The entities that are tagged and relations extracted from the previous steps are combined in this step. The text is read again, and the entities tagged as either subject or object or entity are taken as two entities, and the relation extracted in the previous step is used to form a triple based on the position of the 2 entities and the predicate. The predicate position being in the middle of the 2 entities:

### G. Triples creation II

Second, a graph is generated from the triples from the previous state to uncover the relationships between entities. The graph is used to also include the triples between different sentences. Centered on the connections between prepositions, more triples are produced, such as 'in,' 'on,' 'at' within the graph between named entities. Eventually, the triplets generated by these are combined to produce a complete list of triples.

*H. Filtering of triples*

To improve the efficiency of the triples, we exclude any triple with a stop word. NLTK stop words are included in the stop words, along with commonly used words like figure or number representing the figure.

*I. Filtering by stemming the triples*

The filtered triples obtained from the previous step is further filtered in the sense that only the triples that have either their subject or object in the keyword entity list are included in the final triple list. Further, the subject or object is also stemmed to get the root word and are checked with the keyword entity list for equality.

*1) Lemmatization of triples:* Finally, the triples are lemmatized so that while creating a knowledge graph, both terms such as 'agent' and 'agents' would be viewed as the same entity.

## V. EVALUATION

The method of NER can be scored well against standard assessment criteria for classification by using performance metrics such as accuracy, precision, recall, and F1, but few attempts have paid attention to how organized human intelligence is. It is difficult, in advance, to accurately represent the relationships that when a human reads a piece of text that could be surfaced in someone's mind. For our Evaluation method, we have taken the sample drawn from the output and treated that as a representative of the output and manually checked for actual relations. All relations that seemed valid were given a score of 1 (True relation), while others that do not seem right were marked 0. Based on our 'gold label' and the 'predicted label', a confusion matrix was created, and an approximate estimate of accuracy, precision, recall, and F1 scores are calculated using 'sklearn classification reports'. One challenge that we faced during this method of evaluation was the appearance of 0 in the confusion matrix, which is inevitable since we are using 0 as an output in the predicted label. For taking the results, we focussed on the weighted average score in the report generated by classification. Weighted average ensures that all labels participate according to the proportion of their appearance in the list of gold and predicted labels, hence eliminating the contribution by zero labels in the result. The performance evaluation metrics have been given in figure 2.

|  | precision | recall | f1-score | support |
|---|---|---|---|---|
| 1 | 1.00 | 0.70 | 0.82 | 66 |
| accuracy |  |  | 0.70 | 66 |
| weighted avg | 1.00 | 0.70 | 0.82 | 66 |

Fig. 2. Performance evaluation metrics

## VI. RESULTS

The evaluation score is decent with an accuracy of 70% and an F1 score of 82%. The basic summary of the pipeline for any user using our system is given below :

**Input:**
  *Text from the textbook.
  *Index/Glossary from textbook.

**Output:**
  *All triples from the chapter of the textbook
  *Knowledge graph: visualization of triples

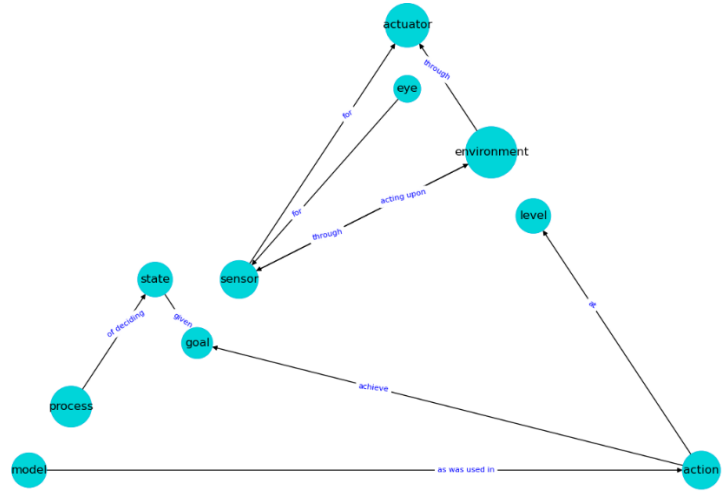

Fig. 3. Knowledge graph for one paragraph in AI textbook

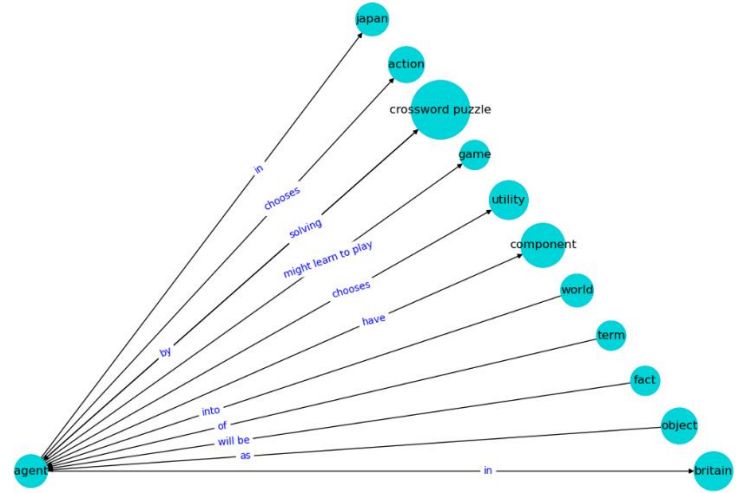

Fig. 4. Knowledge graph for one term 'agent' given in AI textbook chapter

We were able to extract the triples from the given text, and we have shown all different kinds of triples extraction viz—sentences, paragraphs, and related to one entity. Figure 3 shows a Knowledge graph for visualization of triples containing one paragraph entities and relation, while figure 4 shows how all entities are linked with the entity 'agent'.

There are few triples that were missed, mainly due to the non-existence of objects in the list of entities that we acquired from our textbook. Although some triples are lost, we have prevented a lot of invalid/redundant triples that could have formed, and then it would become difficult to filter out the relevant triples.

Another experiment tried out by us included the addition of context to the existing triples that's purely from the index. In this experiment, we tried to add context by including the triple that comes right after the existing triple if the subject of the next triple and the object of the existing triple is the same. But this resulted in many incorrect triples that reduced the recall of the system. Hence there have to be more sophisticated methods for the addition of the context to the main triples.

### A. Limitations

**Ambiguousness in triple**: One big challenge that we faced was with some of the relations such as 'does", and an entity such as 'table'. 'Table' being an entity that was mentioned in the text as an example as well as to show a table representing something. Due to this ambiguity, the triple that was generated was incorrect and hence given predicted/evaluation score 'zero' while doing the human evaluation.

**visualization of triples**: We faced challenges while visualizing the knowledge graph. Due to so many different parameters associated with networkx, out of which many are not documented, we faced difficulties in proper representation of the knowledge graph. Many entities and relations were overlapping while we tried to represent it. After multiple iterations, we were able to figure out the required parameters that worked. However, if we have a large text and many triples since the canvas has a fixed size, the knowledge graph may not be clear enough

## VII. FUTURE WORK

Some work can be done to improve the existing entity recognition system. As of now, we have taken entities from the index, and the triples are generated based on how two entities are related. In doing so, some adjectives related to any triple could be lost. For example, in the sentence 'AI is the future'- AI and future are entities with 'is' as the relation between them, but since the term 'future' is not available in the list of entities from the index, this triple will be lost.

In the approach used in our system, we display only the triples that have either the subject or object in the keyword index. However, for the inclusion of the context in which the triple occurs, we need more sophisticated algorithms to filter the neighboring triples that provide more details or occur in the context of the existing triple.

### A. Competing approaches

One of the competing approaches involves using a semi-supervised learning model, like [10], which works starting with a handful of labeled training examples of each category or relation, plus hundreds of unlabeled data for tagging. This can be used as a substitute for handcrafted methods, as used in our paper.

Another potential approach involves using unsupervised methods like graph-based approaches, as used in [11]. The proposed approach does not require any seed patterns or examples. Instead, it depends on redundancy in large data sets and graph-based mutual reinforcement to induce generalized "extraction patterns".

## VIII. CONCLUSION

With SpaCy, entities were extracted, while POS tags were found and noun and verb phrases were chunks of predefined rules. To extract relations, the entities were mapped into pairs by extracting verbs, prepositions, and post-positions as relation terms. In all, we followed the pipeline-based approach for extracting the triples from any given textbook. Using this method, one can find and extract the majority of the knowledge present in a textbook, irrespective of existing knowledge in the domain. We have also set up a method to visualize the triples extracted in the form of a knowledge graph.

## IX. REFERENCES

[1] C. Xiong, R. Power, and J. Callan, "Explicit semantic ranking for academic search via knowledge graph embedding," in Proc. WWW, 2017, pp. 1271–1279.
[2] Y. Zhang, H. Dai, Z. Kozareva, A. J. Smola, and L. Song, "Variational reasoning for question answering with knowledge graph," in Proc. AAAI, 2018.
[3] Z. Sun, J. Yang, J. Zhang, A. Bozzon, L.-K. Huang, and C. Xu, "Recurrent knowledge graph embedding for effective recommendation," in Proc. ACM RecSys, 2018, pp. 297–305.
[4] D. Vrandecic and M. Krotzsch, "Wikidata: a free collaborative knowledge base," Communications of the Acm, vol. 57, no. 10, pp. 78–85, 2014.
[5] Q. Liu, Y. Li, H. Duan, Y. Liu, and Z. Qin, "Knowledge graph con- struction techniques," Journal of Computer Research and Development, vol. 53, no. 3, pp. 582–600, 2016.
[6] H. Paulheim, "Knowledge graph refinement: A survey of approaches and evaluation methods," Semantic web, vol. 8, no. 3, pp. 489–508, 2017.
[7] I. Augenstein, M. Das, S. Riedel, L. Vikraman, and A. McCallum, "Semeval 2017 task 10: Science-extracting keyphrases and relations from scientific publications," arXiv preprint arXiv:1704.02853, 2017.
[8] https://www.groundai.com/project/unsupervised-construction-of-knowledge-graphs-from-text-and-code/1
[9] Stewart, Michael, Majigsuren Enkhsaikhan, and Wei Liu. "Icdm 2019 knowledge graph contest: Team uwa." In 2019 IEEE International Conference on Data Mining (ICDM), pp. 1546-1551. IEEE, 2019.
[10] Carlson, Andrew, Justin Betteridge, Richard C. Wang, Estevam R. Hr- uschka Jr, and Tom M. Mitchell. "Coupled semi-supervised learning for information extraction." In Proceedings of the third ACM international conference on Web search and data mining, pp. 101-110. 2010.
[11] Mihalcea, Rada, and Dragomir Radev. Graph-based natural language processing and information retrieval. Cambridge university press, 2011.
[12] Ratner, Alexander, Stephen H. Bach, Henry Ehrenberg, Jason Fries, Sen Wu, and Christopher Ré. "Snorkel: Rapid training data creation with weak supervision." In Proceedings of the VLDB Endowment. International Conference on Very Large Data Bases, vol. 11, no. 3, p.
269. NIH Public Access, 2017.
[13] Artificial Intelligence, A Modern Approach (Third Edition) by Stuart Russell and Peter Norvig.